\documentclass[sigconf, nonacm]{acmart}

\AtBeginDocument{%
  }

\setcopyright{acmlicensed}
\copyrightyear{2026}
\acmYear{2026}
\acmConference[Conference acronym 'SIGIR]{The 49th International
ACM SIGIR Conference on Research and Development in Information Retrieval
(SIGIR 2026). ACM, }{20–24 July,
  2026}{Melbourne | Naarm, Australia}

\usepackage{amsthm, bbm, bm, dsfont, enumitem}
\usepackage{CJKutf8}  
\usepackage{setspace, comment, algorithm, latexsym}
\usepackage[flushleft]{threeparttable}
\usepackage{longtable}
\allowdisplaybreaks
\usepackage{algpseudocode}
\usepackage{wrapfig,lipsum}
\usepackage{makecell}
\usepackage{CJKutf8}
\usepackage{multirow}
\usepackage{booktabs}
\usepackage{multirow}
\usepackage{makecell}
\theoremstyle{plain}

\usepackage{url}            
\usepackage{nicefrac}       
\usepackage{subcaption}
\usepackage{adjustbox}

\begin{document}

\title{QUARK: Robust Retrieval under Non-Faithful Queries via Query-Anchored Aggregation} 

\author{Rita Qiuran Lyu}
\affiliation{%
  \institution{University of California, Berkeley}
  \city{Berkeley}
  \state{CA}
  \country{USA}
}
\email{qiuran\_lyu@berkeley.edu}

\author{Michelle Manqiao Wang}
\affiliation{%
  \institution{Carnegie Mellon University}
  \city{Pittsburgh}
  \state{PA}
  \country{USA}}
\email{manqiaow@andrew.cmu.edu}

\author{Lei Shi}
\affiliation{%
  \institution{Adobe Research}
  \city{San Jose}
  \state{CA}
  \country{USA}
}
\email{leis@adobe.com}

\renewcommand{\shortauthors}{Trovato et al.}

\begin{abstract}
User queries in real-world retrieval are often \emph{non-faithful}—noisy, incomplete, or distorted—causing retrievers to fail when key semantics are missing. We formalize this as \emph{retrieval under recall noise}, where the observed query is drawn from a noisy recall process of a latent target item. To address this, we propose QUARK, a simple yet effective training-free framework for robust retrieval under non-faithful queries. QUARK explicitly models query uncertainty through recovery hypotheses—multiple plausible interpretations of the latent intent given the observed query—and introduces query-anchored aggregation to combine their signals robustly. The original query serves as a semantic anchor, while recovery hypotheses provide controlled auxiliary evidence, preventing semantic drift and hypothesis hijacking. This design enables QUARK to improve recall and ranking quality without sacrificing robustness, even when some hypotheses are noisy or uninformative. Across controlled simulations and BEIR benchmarks (FIQA, SciFact, NFCorpus) with both sparse and dense retrievers, QUARK improves Recall, MRR, and nDCG over the base retriever. Ablations show QUARK is robust to the number of recovery hypotheses and that anchored aggregation outperforms unanchored max/mean/median pooling. These results demonstrate that modeling query uncertainty through recovery hypotheses, coupled with principled anchored aggregation, is essential for robust retrieval under non-faithful queries.
\end{abstract}



\received{20 February 2007}
\received[revised]{12 March 2009}
\received[accepted]{5 June 2009}

\maketitle

\section{Introduction}\label{sec:introduction}

Modern information retrieval (IR) systems are usually designed under an implicit assumption that user queries can faithfully express an underlying information need. However, in many scenarios, this assumption does not hold. Users may issue queries that are incomplete, noisy, paraphrased, or distorted due to memory limitations, ambiguity, or language mismatch \citep{krovetz1992lexical,jansen1998real, zhang2020query}. We refer to such inputs as \emph{non-faithful queries}, where the observed query is an imperfect manifestation of a latent user intent.

Retrieval under non-faithful queries is challenging because even strong retrievers can fail when critical terms are missing or incorrectly expressed \citep{furnas1987vocabulary,belkin1980anomalous}. While more powerful retrievers improve robustness to lexical variation, they remain sensitive to semantic drift and often exhibit unstable behavior under severe query noise \citep{thakur2021beir, liu2024perturbation}. Moreover, users often reformulate queries unconsciously \citep{chen2021towards, jansen2009patterns}, indicating that a single-shot retrieval is not sufficient.

One way for handling imperfect queries is to \emph{rewrite} them via expansion, correction, or paraphrasing, so that retrieval is performed on a “cleaner” formulation \citep{azad2019query, elgohary2019can, qian2022explicit}. However, under non-faithful queries, rewriting is also uncertain: one noisy query may admit multiple plausible interpretations, and committing to a single rewrite can be brittle when ambiguity is high or key evidence is missing.

In this work, we propose \textbf{QUARK}, a simple yet effective training-free framework for robust retrieval under non-faithful queries.  QUARK generates a set of \emph{recovery hypotheses} based on the observed query and aggregates their retrieval signals using a \emph{query-anchored aggregation} strategy. The original query serves as an anchor that stabilizes ranking, while recovery hypotheses contribute auxiliary evidence in a controlled manner. QUARK is also retriever-agnostic and can be applied on top of existing retrieval systems.

We conduct extensive experiments on both controlled simulation settings and real-world benchmarks, demonstrating that QUARK can improve retrieval quality without sacrificing robustness. Our ablation studies show that QUARK is insensitive to the exact number of recovery hypotheses and that query-anchored aggregation is crucial for avoiding performance collapse. These results suggest that robust retrieval under non-faithful queries requires not only generating alternative hypotheses, but also aggregating them in a principled, intent-preserving way.

In summary, the main contributions of this paper are as follows:
\begin{itemize}
    \item We formalize retrieval under non-faithful queries as a latent-intent inference problem, where the observed query is a noisy, partial realization of an unobserved user intent. Our formulation disentangles (i) the latent target item, (ii) the query corruption process that generates the issued query, and (iii) an optional recovery mechanism that produces candidate query hypotheses. This modeling view provides a principled foundation for analyzing robustness and designing retrieval methods for imperfect queries.
    \item We propose QUARK, a training-free retrieval framework that generates multiple recovery hypotheses and aggregates their retrieval signals through a query-anchored weighting mechanism. QUARK preserves the semantic anchor of the original query while selectively incorporating evidence from recovery hypotheses, effectively mitigating semantic drift and score dilution.
    \item Extensive experiments on controlled simulation settings and multiple real-world benchmarks demonstrate that QUARK consistently improves Recall@M, MRR@M, and nDCG@M over strong lexical and dense retrievers. Notably, QUARK never degrades performance across different noise levels and retriever backbones, highlighting its robustness and practical applicability.
\end{itemize}

\section{Related Work}
\label{sec:related work}

\subsection{Ad-hoc Information Retrieval}
Ad-hoc IR systems return relevant documents for short, standalone queries without prior context. Classical lexical approaches such as BM25 \citep{robertson1995okapi} and TF-IDF \citep{salton1988term} remain strong baselines due to their efficiency and interpretability. More recently, dense retrieval methods based on encoders and pretrained language models have achieved improvements by capturing semantic similarity beyond surface forms. This includes LLM2Vec~\citep{behnamghader2024llm2vec}, {DPR}~\cite{karpukhin2020dense}, {ANCE}~\cite{xiong2020approximate}, and late-interaction models like {ColBERT}~\cite{khattab2020colbert}.

Despite these advances, both lexical and dense retrievers remain vulnerable to query imperfections. Lexical methods degrade sharply when key terms are missing or distorted~\cite{thakur2021beir}, while dense models can suffer from semantic overgeneralization or spurious similarity under noisy inputs~\cite{liu2025robust}. These limitations motivate complementary strategies that improve robustness without retraining retrievers, which is where QUARK is positioned.

\subsection{Query Expansion and Reformulation}
Query expansion and reformulation aim to improve retrieval by augmenting the original query with additional terms or alternative expressions. Previous work for automatic query expansion includes knowledge-based generation \citep{bodner1996knowledge}, pseudo-relevance-based based generation \citep{azad2022improving,wang2023generative}, and so on \citep{carpineto2012survey,afuan2019study, azad2019query}. Neural methods generate paraphrases or expansions using sequence-to-sequence models~\cite{nogueira2019multi,wang2024utilizing}. More recently, large language models  (LLMs) have been applied to query rewriting~\cite{sun2024r,liu2024query}, paraphrasing~\citep{krishna2023paraphrasing} and intent clarification~\citep{aliannejadi2019asking, zhu2025large}. While expansion-based methods can improve recall, they are still sensitive to expansion quality and often introduce noise if expansions drift away from the true intent. 

\subsection{Robust Retrieval under Query Noise}
Robust retrieval \citep{liu2025robust} has been studied in the context of noisy, ambiguous, or adversarial queries, including spelling errors~\cite{sidiropoulos2022analysing}, paraphrasing~\cite{gan2019improving,agarwal2023towards}, and underspecified inputs. Some approaches focus on improving retriever robustness through  adversarial training~\cite{liu2025robust,zhou2023towards}, while others rely on user interaction or iterative refinement~\cite{jang2024itercqr,zhou2014iterative}. Character-level models and phonetic encoding have also been explored to handle surface-form variations~\cite{ng1998phonetic,chen2018phonetic}.

Our work differs in that we target robustness at inference time without retriever retraining. 

\subsection{Fusion and Aggregation Methods}

Fusion and aggregation methods combine evidence from multiple retrieval signals, such as rankings, scores, or query variants, to improve overall retrieval effectiveness.
Early work in IR studied rank fusion techniques that aggregate multiple ranked lists using heuristics such as CombSUM, CombMNZ, and Reciprocal Rank Fusion (RRF) \cite{fox1993combining, cormack2009reciprocal}.
These methods are simple, retriever-agnostic, and often robust in practice, but they treat all input signals symmetrically and do not account for the reliability or intent alignment of individual inputs.

More recent work has explored score-level and representation-level fusion in neural retrieval.
Examples include combining lexical and dense retrievers \cite{song2025sparse}, late interaction models \cite{khattab2020colbert}, and hybrid systems that merge sparse and dense representations \cite{kong2023sparseembed}.
While effective, these approaches typically assume that each retrieval signal corresponds to a faithful or complementary view of the same query, and they do not explicitly model uncertainty arising from imperfect or noisy query formulations.

In contrast, our work treats recovery hypotheses as uncertain auxiliary evidence generated from a non-faithful query, rather than as interchangeable rewrites.
QUARK adopts a query-anchored aggregation mechanism that conditions aggregation on the original query, selectively incorporating hypothesis-level signals while suppressing misaligned evidence.
This distinguishes our approach from prior unanchored fusion methods and enables robust retrieval improvements even when some hypotheses are incorrect.

\section{Problem Setup}
\label{sec:setup}

Before presenting our method, we first introduce a formal problem setup for retrieval under \emph{non-faithful queries}, where ``non-faithful'' means that the observed query is an imperfect expression of an unobserved user intent.

\subsection{Retrieval with Latent User Intent}
\label{sec:setup_intent}

Let $\mathcal{D}=\{d_1,\dots,d_N\}$ denote an indexed corpus of $N$ retrievable items (e.g., documents, passages, or entities).
For each retrieval interaction (i.e., each issued query), a user has an unobserved information need denoted by a latent intent variable $Z \in \{1,\dots,N\}$, which indexes the target item $d^\star := d_Z \in \mathcal{D}$.
However, the retrieval system cannot observe $Z$ or $d^\star$ directly. It only receives an observed query string $q$ (a sequence of tokens from a vocabulary $\Sigma$).

For example, in web search, $\mathcal{D}$ is the engine's indexed pages. A user query such as ``apple founder stanford speech'' may implicitly target a specific page $d^\star$ containing the desired talk, while the system only observes the typed query text $q$.

\subsection{Query Observation via a Recall Channel}
\label{sec:setup_channel}
In real-world settings, the observed query $q$ may be either a faithful or a non-faithful expression of the target item $d^\star$.
We model $q$ as an observation generated from $d^\star$ through an unknown \emph{recall process},
\begin{equation}
q \sim P_{\mathrm{recall}}(\,\cdot \mid d^\star),
\label{eq:recall_channel}
\end{equation}
where the conditional distribution $P_{\mathrm{recall}}(\cdot \mid d^\star)$ is unobserved and is not assumed to preserve lexical or semantic similarity.
Equivalently, one may write $q = T(d^\star;\varepsilon)$, where $\varepsilon$ denotes latent noise factors and $T(\cdot)$ is a stochastic transformation that maps the target item to the observed query.

\subsubsection{Sources of Noise}
\label{sec:setup_channel_sources}

The noise factor $\varepsilon$ in the recall channel may introduce heterogeneous distortions, including but not limited to:
(i) surface-form corruption (e.g., substitutions, omissions, segmentation errors),
(ii) phonetic or orthographic confusion,
and (iii) semantic drift due to paraphrasing or abstraction.
These noises can substantially reduce the effectiveness of standard lexical or embedding-based retrieval methods.

\subsubsection{Faithfulness and Non-faithfulness}
\label{sec:setup_faithfulness}

The recall channel in Eq.~\eqref{eq:recall_channel} allows the observed query $q$ to deviate from the target item $d^\star$ in surface form and/or meaning. 
For example, in web search for Steve Jobs' Stanford talk, a faithful query may closely resemble the target text (e.g., ``Steve Jobs Stanford commencement speech''), whereas non-faithful queries may exhibit surface-form corruption (e.g., ``Jobs Standhord speach''), phonetic or orthographic confusion (e.g., ``apple founder stanferd talk''), or semantic abstraction (e.g., ``three stories from a famous Stanford graduation''). 
Although they refer to the same target item, these queries can differ substantially in surface form and semantics.

In controlled settings where the ground-truth target $d^\star$ is available (e.g., curated benchmarks or simulations), we can quantify how faithful a query is to its target via a score
\begin{equation}
\phi(q,d^\star) \in \mathbb{R},
\end{equation}
where larger values indicate higher faithfulness under a chosen notion of similarity. Importantly, $\phi(q,d^\star)$ is used only for analysis and controlled evaluation, and is \emph{not} assumed to be observable by the retrieval system at inference time.


\subsection{Retrieval Objective}
\label{sec:setup_objective}

Given an observed query $q$, the retrieval system produces a ranked list over the corpus $\mathcal{D}$ using a scoring function $S(q,d)$ for $d \in \mathcal{D}$.
Let $\pi(q)$ denote the induced ranking, and let $\pi_M(q)$ be the top-$M$ retrieved items.
The retrieval objective is to rank the latent target item $d^\star$ as highly as possible. In evaluation, we report standard ranking metrics including Recall@$M$, MRR@$M$, and nDCG@$M$, and analyze performance across different query faithfulness regimes.

\subsection{Methodological Implications}
\label{sec:setup_implication}

Under the recall-channel view, retrieval can be conceptually interpreted as inferring the latent intent $Z$ from an observed query $q$, leading to a posterior of the form
\begin{equation}
P(Z \mid q) \propto P_{\mathrm{recall}}(q \mid d_Z)\,P(Z),
\end{equation}
where the proportionality reflects normalization over all candidate items.
In practice, we do not explicitly model the prior $P(Z)$ and assume no additional intent-specific information beyond what is captured by the retrieval process.

A key challenge arises from the fact that the recall distribution $P_{\mathrm{recall}}$ is unknown and may induce substantial distortions between the target item and the observed query.
Under non-faithful queries, this often results in a diffuse posterior over candidate intents, where many items appear similarly plausible.
As a result, retrieval methods that rely on a single query similarity signal may fail to robustly recover the latent target item.


\section{Methodology}
\label{sec:method}
\subsection{Overview of the Proposed Framework}
\label{sec:method_overview}

Under the formulation in Section~\ref{sec:setup}, retrieval under non-faithful queries is challenging because the observed query $q$ may provide an incomplete or distorted view of the user's latent intent.
As a result, relying on a single representation of the original query $q$ can lead to brittle retrieval behavior, especially when recall noise substantially alters surface form or semantics.
To address this challenge, we propose \textsc{QUARK} (
\textbf{Q}uery-anchored \textbf{U}ncertainty-aware \textbf{A}ggregation for \textbf{R}etrieval over top-\textbf{K} interpretations), a \emph{training-free} framework that explicitly accounts for uncertainty in query interpretation by aggregating evidence across multiple plausible reconstructions of the latent intent. 

\begin{figure}[t]
    \centering
    \includegraphics[width=\linewidth]{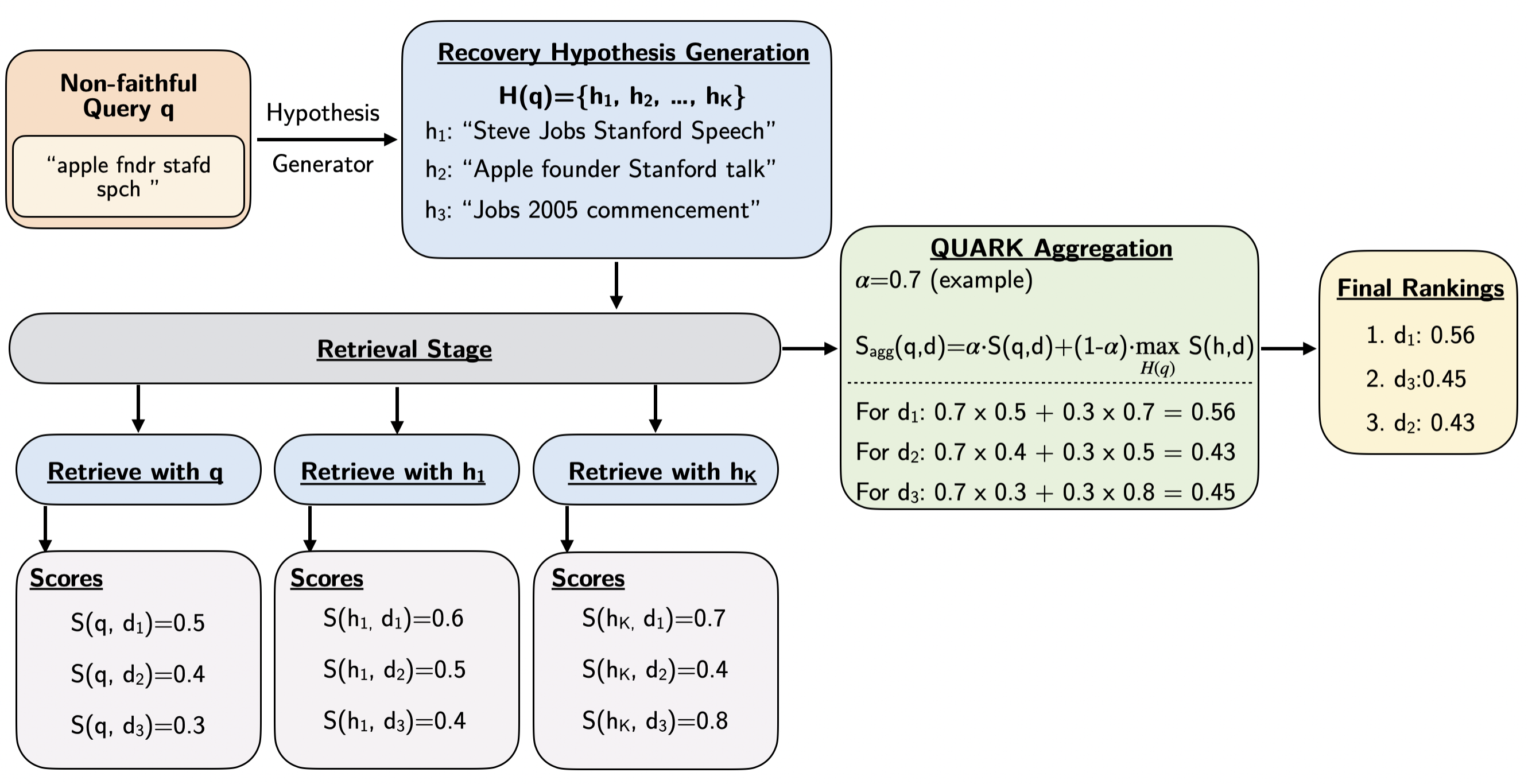}
    \caption{The overview of QUARK.}
    \label{fig:workflow}
\end{figure}

Given an observed query $q$, \textsc{QUARK} first generates a set of \emph{recovery hypotheses}, each representing a plausible alternative expression of the user's intended target under recall noise.
These hypotheses are not assumed to be correct individually; rather, they provide complementary views that collectively cover different ways the latent intent may be expressed.
Each hypothesis is used to retrieve candidate items from the corpus, producing multiple hypothesis-conditioned rankings.
The final retrieval result is obtained by aggregating these rankings in a manner anchored to the original query, ensuring stable evidence integration without allowing any single hypothesis to dominate (Algorithm~\ref{alg:qar}).
Because QUARK operates purely at inference time, it can be applied on top of any existing retriever without additional training or index modification. The overview of QUARK is presented in Figure~\ref{fig:workflow}.

\subsection{Recovery Hypothesis Generation}
\label{sec:method_hypothesis}
Instead of deterministically ``correcting'' $q$, we generate \emph{recovery hypotheses} to represent multiple plausible interpretations of an observed query $q$ under recall noise. This can expand the space of candidate expressions that may align with the user's latent intent, providing complementary views of what the user could have meant. 

Formally, given $q$, we produce a set of hypotheses $\mathcal{H}(q)=\{h_1,\dots,h_K\}$, where each $h_k$ is an alternative expression conditioned on $q$ to cover diverse plausible interpretations under recall noise. The hypotheses are constructed to be diverse yet relevant in order to capture lexical, phonetic, and semantic cues that may be altered by the unknown recall channel. Importantly, a hypothesis need not preserve surface similarity to $q$ nor exactly match the ground-truth target item $d^\star$; its role is to contribute evidence toward the latent intent when used as a query in downstream retrieval.

\subsection{Query-Anchored Aggregation}
\label{sec:method_aggregation}

Let $S(\cdot,\cdot)$ denote a base retrieval scoring function that assigns a relevance score to an item $d \in \mathcal{D}$ given a query input.
In addition to the original observed query $q$, we consider a set of recovery hypotheses $\mathcal{H}(q)=\{h_1,\dots,h_K\}$.
For each input (either $q$ or $h_k$), we compute retrieval scores $S(q,d)$ and $S(h_k,d)$, respectively.

We aggregate retrieval evidence using a single anchoring parameter $\alpha \in [0,1]$ that controls the contribution of the original query relative to the recovery hypotheses.

Concretely, for each item $d$, we form an aggregated score by interpolating between the score induced by the original query and the strongest score induced by any recovery hypothesis:
\begin{equation}
S_{\mathrm{agg}}(q,d)
\;=\;
\alpha \, S(q,d)
\;+\;
(1-\alpha)\,\max_{h \in \mathcal{H}(q)} S(h,d),
\label{eq:agg_score}
\end{equation}
and rank items according to $S_{\mathrm{agg}}(q,d)$. When $\alpha=1$, Eq.~\eqref{eq:agg_score} reduces to standard single-query retrieval using $q$ alone.
When $\alpha=0$, the ranking is determined entirely by the recovery hypotheses (excluding the original query), using the strongest hypothesis support for each item via the max operator.
Intermediate values interpolate between these extremes, enabling a controlled trade-off between anchoring to the observed query and leveraging alternative interpretations.
This max-based aggregation emphasizes the most supportive hypothesis for each item while remaining anchored to the original query, improving robustness under non-faithful query observations without relying on a single query representation. We present the whole algorithm in Algorithm ~\ref{alg:qar}. 

\begin{algorithm}[t]
\caption{Query-Anchored Retrieval under Non-Faithful Queries}
\label{alg:qar}
\begin{algorithmic}
\Require Observed query $q$, corpus $\mathcal{D}$, retrieval scoring function $S(\cdot,\cdot)$, anchoring parameter $\alpha \in [0,1]$, number of hypotheses $K$
\Ensure Ranked list of items from $\mathcal{D}$

\Statex \textbf{Step 1: Recovery Hypothesis Generation}
\State Generate recovery hypothesis set $\mathcal{H}(q)=\{h_1,\dots,h_K\}$ conditioned on $q$

\Statex \textbf{Step 2: Hypothesis-Conditioned Retrieval}
\State Compute retrieval scores $S(q,d)$ for all $d \in \mathcal{D}$
\For{each hypothesis $h_k \in \mathcal{H}(q)$}
    \State Compute retrieval scores $S(h_k,d)$ for all $d \in \mathcal{D}$
\EndFor

\Statex \textbf{Step 3: Query-Anchored Aggregation}
\For{each item $d \in \mathcal{D}$}
    \State Compute aggregated score:
    \[
    S_{\mathrm{agg}}(q,d)
    =
    \alpha \, S(q,d)
    +
    (1-\alpha)\max_{h \in \mathcal{H}(q)} S(h,d)
    \]
\EndFor

\State Rank items in $\mathcal{D}$ by $S_{\mathrm{agg}}(q,d)$ in descending order

\noindent\Return Ranked list
\end{algorithmic}
\end{algorithm}

\subsection{Illustrative Example: Preventing Semantic Drift By Query-Anchoring}
\label{sec:method_example}
In this subsection, we illustrate why \emph{query-anchored} aggregation is necessary in QUARK to prevent \emph{semantic drift}. While recovery hypotheses can provide complementary evidence under recall noise, aggregating them without additional structure can also introduce a new failure mode: the final ranking may drift toward an interpretation that is plausible but not intended.

Because hypotheses explore alternative interpretations of the observed query $q$, some may capture tangential or overly abstract aspects that deviate from the user's intent.
We refer to the resulting failure mode as \emph{hypothesis hijacking}: a drifting hypothesis that attains a high retrieval score $S(h,d)$ can dominate aggregation and steer the ranking toward items that are plausible under that hypothesis but misaligned with the true target $d^\star$.

The anchoring parameter $\alpha$ in Eq.~\eqref{eq:agg_score}, therefore, mitigates hypothesis hijacking by retaining a weighted contribution from the original query.
In particular, $\alpha=1$ recovers standard single-query retrieval, whereas smaller $\alpha$ increases reliance on hypotheses in a controlled manner, improving robustness under non-faithful queries without uncontrolled drift.

\subsection{Interpretation as Monte Carlo Aggregation}
\label{sec:method_mc}

Finally, we provide a probabilistic interpretation of the aggregation procedure in QUARK under the recall-channel view introduced in Section~\ref{sec:setup}.
Recall that retrieval under recall noise can be conceptually viewed as inferring the latent intent $Z$ from an observed query $q$, where the uncertainty arises from the unknown recall distribution $P_{\mathrm{recall}}(q \mid d_Z)$.
In this setting, the recovery hypotheses $\mathcal{H}(q)$ can be interpreted as samples from an implicit proposal mechanism that explores plausible realizations of the user's intent under recall noise. Now, aggregating retrieval evidence across hypotheses implicitly approximates a marginalization over uncertain query interpretations and  the aggregated score in Eq.~\eqref{eq:agg_score} can be viewed as a Monte Carlo–style approximation to intent-level evidence accumulation, where the contribution of the original query is explicitly preserved and hypothesis-induced evidence is incorporated through a robust estimator.
The $\max$ operator selects the strongest hypothesis per document, making aggregation robust to noisy or uninformative hypotheses in $\mathcal{H}(q)$. This ensures each document's score reflects its best alignment with any plausible query interpretation.

\section{Experiments}\label{sec:experiments}

\subsection{Datasets and Evaluation Metrics}
\label{sec:exp_setup}

We evaluate QUARK in both a controlled lyrics-retrieval simulation study and three real-world retrieval benchmarks.
The simulation setting allows precise control over query faithfulness and recall noise, while the real datasets test robustness under possibly non-faithful queries in real world.

\subsubsection{Lyrics-retrieval simulation.}
We build a Chinese lyrics-retrieval benchmark designed to model querying behavior with different level of recall noise. The corpus $\mathcal{D}$ consists of 2,179,443 Chinese lyric units drawn from 49,761 songs by 494 artists crawled from NetEase Cloud Music \citep{gaussic_chinese_lyric_corpus}, where each unit is a unique retrievable item. We create 5,000 seed queries by (i) filtering lyric units by simple heuristics (length 8–50 characters; remove interjection-only lines), (ii) grouping units by song, and (iii) randomly sampling up to 3 units per song until reaching 5,000. Each sampled unit is treated as a ground-truth target $d^\star$. 

We use a pretrained LLM  \texttt{Qwen3-max} \citep{yang2025qwen3}  to generate observed queries $q$ under different noise levels  conditioned on the seed queries. This process reflects realistic user behavior in which queries are generated from memory rather than copied verbatim from the target text. To be specific, we generate queries under three prompting regimes that induce different levels of query non-faithfulness:
\emph{L1: mildly non-faithful}, \emph{L2: moderately non-faithful}, and \emph{L3: severely non-faithful}.
These regimes differ in the degree of lexical distortion, paraphrasing, and semantic abstraction allowed during query generation. The prompts used are  listed in Figure ~\ref{fig:prompt_levels}.

\begin{figure*}[t]
\fbox{
\parbox{0.97\linewidth}{
\small
\textbf{Prompt Templates for Lyric-Recall Query Generation (L1--L3).}

\textbf{L1: Mildly Non-Faithful Recall}

You are a listener of Chinese songs. You will be shown one line of correct lyrics.\\
Please generate a \emph{lightly distorted} recalled version.

\textbf{Allowed changes (at most two total):} replace one character with a homophone or near-homophone; replace one word with a near-synonym; optionally omit one function word or slightly adjust punctuation.\\
\textbf{Requirements:} sentence structure remains mostly unchanged; overall meaning remains almost identical; output reads like a natural lyric line.\\
\textbf{Forbidden:} major rewriting or expansion; introducing new themes or scenes.\\
\textbf{Output:} only the recalled lyric line. No explanation.\\
\textbf{Original lyric:} \texttt{"\{lyric\}"}.

\textbf{L2: Moderately Non-Faithful Recall}

You are a listener of Chinese songs. You will be shown one line of correct lyrics.\\
Please generate a \emph{moderately distorted} recalled version.

\textbf{Allowed changes:} make 2--4 homophone/near-homophone substitutions or segmentation/character-selection errors; replace 1--2 words with more common but not fully correct collocations; slightly adjust local word order while keeping sentence length and rhythm roughly similar.\\
\textbf{Requirements:} emotional tone or imagery remains recognizable; the recalled version is not a continuous substring of the original lyric.\\
\textbf{Forbidden:} completely changing the theme; turning the line into explanatory prose.\\
\textbf{Output:} only the recalled lyric line. No explanation.\\
\textbf{Original lyric:} \texttt{"\{lyric\}"}.

\textbf{L3: Severely Non-Faithful Recall}

You are a listener of Chinese songs. You will be shown one line of correct lyrics.\\
Please generate a \emph{severely distorted} recalled version.

\textbf{Recall setting:} most exact wording is forgotten; retain 1--2 imagery or emotional cues from the original lyric; remaining content may be rewritten; vague references are allowed.\\
\textbf{Requirements:} do not preserve original structure or continuous fragments; emotional tone or scene remains similar; output still reads like a lyric line.\\
\textbf{Forbidden:} explanatory prose; removing all original lyric cues; introducing unrelated themes.\\
\textbf{Output:} only the recalled lyric line. No explanation.\\
\textbf{Original lyric:} \texttt{"\{lyric\}"}.

}}
\caption{Prompt templates used with \texttt{Qwen3-max} \citep{yang2025qwen3} to generate observed lyric-recall queries under three levels of non-faithfulness in the lyrics-retrieval simulation.}
\label{fig:prompt_levels}
\end{figure*}

To validate the effectiveness of different prompts, we quantify faithfulness post hoc using character-level ROUGE-L F1 (R-L$_c$), normalized edit similarity (EditSim), and longest common substring length (LCS) between the generated query $q$ and the target lyric $d^\star$, where EditSim$(q,d^\star)=1-\frac{\mathrm{ED}(q,d^\star)}{\max(|q|,|d^\star|)}$, 
and $\mathrm{ED}(\cdot,\cdot)$ is the Levenshtein (edit) distance. These metrics can serve as the score $\phi(q,d^\star)$ defined in Section~\ref{sec:setup_faithfulness}. In Table~\ref{tab:faithfulness_stats_twocol}, we observe clear separation across the three regimes, validating that the prompts induce distinct recall-noise conditions.

\begin{table}[t]
\centering

\setlength{\tabcolsep}{3pt}  
\begin{tabular}{llccccc}
\toprule
\textbf{Recall Level} & \textbf{Metric} &
\textbf{Mean} & \textbf{Med} & \textbf{Std} & \textbf{Min} & \textbf{Max} \\
\midrule

\multirow{5}{*}{L1} 
& R-L$_c$ & 0.900 & 0.900 & 0.078 & 0 & 1\\
& EditSim & 0.881 & 0.889 & 0.089 & 0 & 1 \\
& LCS & 7.380 & 7.000 & 2.410 & 0 & 32 \\
& $|q|$ & 9.640 & 9.000 & 2.100 & 2 & 34 \\
& $|d^\star|$ & 9.760 & 9.000 & 2.050 & 8 & 32 \\
\midrule

\multirow{5}{*}{L2} 
& R-L$_c$ & 0.783 & 0.800 & 0.136 & 0 & 1 \\
& EditSim & 0.754 & 0.778 & 0.154 & 0 & 1 \\
& LCS & 5.420 & 5 & 2.370 & 0 & 32 \\
& $|q|$ & 9.930 & 9 & 2.200 & 2 & 36 \\
& $|d^\star|$ & 9.760 & 9.000 & 2.050 & 8 & 32 \\
\midrule

\multirow{5}{*}{L3} 
& R-L$_c$ & 0.414 & 0.417 & 0.156 & 0 & 0.950 \\
& EditSim & 0.269 & 0.250 & 0.178 &0 & 0.910 \\
& LCS & 2.260 & 20 & 1.140 & 0 & 19 \\
& $|q|$ & 10.110 & 10 & 2.130 & 5 & 25 \\
& $|d^\star|$ & 9.760 & 9& 2.050 & 8 & 32 \\
\bottomrule
\end{tabular}

\caption{Post-hoc faithfulness statistics across recall levels. 
R-L$_c$ = ROUGE-L at character level; LCS = longest common substring length; $|q|$ and $|d^\star|$ are the length of observed query and gold target.}
\label{tab:faithfulness_stats_twocol}
\end{table}

\subsubsection{Real-world retrieval benchmarks.}
We further test QUARK on three public retrieval benchmarks from the BEIR collection \citep{thakur2021beir}: {SciFact}, {NFCorpus}, and {FiQA}. Unlike the simulation setting where $d^\star$ is known and recall noise is synthetically controlled, queries in real-world benchmarks may already exhibit unknown degrees of surface-form corruption, paraphrasing, or semantic abstraction. 

To be specific, SciFact focuses on scientific claim verification, where short factual queries must retrieve relevant scientific abstracts. Queries are often concise and may only partially reflect the target document wording.  NFCorpus contains health-related search queries posed by real users, which frequently include lay terminology, paraphrasing, or incomplete symptom descriptions.  
FiQA is a financial question answering benchmark, where queries are natural-language questions over financial news and reports, often including domain-specific phrasing or abstraction.

\subsubsection{Evaluation Metrics}
Across both settings, we  report Recall@$M$ (R@$M$), MRR@$M$, and nDCG@$M$ to evaluate overall ranking quality.

Statistical significance is evaluated only for MRR and nDCG, which are continuous, rank-sensitive metrics that vary smoothly across queries and are therefore amenable to paired statistical testing. We use a two-sided paired $t$-test over queries to compare QUARK with the corresponding base retriever. We do not perform statistical tests for recall-based metrics (Recall@1/5/10), which are discrete and thresholded, and thus less suitable for parametric significance testing.

\subsection{Implementation of Recovery Hypothesis Generation}
\label{subsec:implementation_hypothesis}

We use \texttt{Qwen-3-Max} \citep{yang2025qwen3} as a hypothesis sampler that maps an observed query to a small set of alternative hypotheses without any task-specific fine-tuning.

\subsubsection{Rationale for Using LLMs.}
The use of LLMs as hypothesis sampler is motivated by the nature of non-faithful queries. In realistic retrieval scenarios, deviations between a user’s latent information need and the observed query arise from complex recall processes, including paraphrasing, partial memory, abstraction, omission, phonetic approximation, and stylistic variation. These processes are heterogeneous, implicit, and difficult to enumerate through rules or lightweight transformation heuristics.

Pretrained language models implicitly capture such recall variability through exposure to large-scale, diverse text corpora, making them well-suited to approximate sampling from an underlying recall distribution. Importantly, within our framework, the LLM is not used to infer the true intent, correct errors, or rank candidates. Its sole role is to generate a set of plausible, intent-preserving query hypotheses, which are later handled by the downstream retriever and aggregation mechanism. 

\subsubsection{Prompt Design Principles.}
To ensure that the generated hypotheses are not artificial perturbations, we follow the following prompt design principles:

\begin{enumerate}
    \item \textbf{Implicit Noise Modeling.}  
    To avoid injecting synthetic structure , the prompt does not reference explicit noise levels, error categories, or faithfulness labels. 
    \item \textbf{General and High-Level Instructions.}  
    To encourage diversity while preserving the latent intent, the prompt provides abstract guidance (e.g., allowing rephrasing, omission, or generalization) rather than specifying concrete edit rules. 

    \item \textbf{Recall-Oriented Generation.}  
    Generated hypotheses are framed as plausible recalled expressions rather than improved, corrected, or canonical rewrites. The model is explicitly instructed not to judge correctness or select among candidates.

    \item \textbf{Separation from Retrieval.}  
    The hypothesis generator operates independently of the retriever. No retrieval signals, ranking objectives, or feedback are incorporated into the generation process.
\end{enumerate}
\subsubsection{Prompt Instantiation Across Datasets.}
We implement recovery hypothesis generation using \texttt{Qwen-3-Max} with lightweight, domain-adapted prompts. Unless otherwise noted, we set $K=5$.
\noindent\textbf{Lyrics Retrieval Simulation.} For the lyrics-retrieval simulation, following the prompt design principles, we prompt the hypothesis generator as a Chinese song search assistant and treat the observed query as a potentially incorrect or incomplete lyric recall.  For reproducibility, we present the full prompt used in Figure~\ref{fig:lyrics_prompt}.
\begin{figure}[t]
\small
\setlength{\fboxsep}{5pt}
\fbox{
\begin{minipage}{\columnwidth}
\textbf{Lyrics Retrieval Prompt.}

You are a Chinese song search assistant.

The user input is a recalled lyric that may be incorrect or incomplete.
The wording, order, or details may deviate from the original lyrics,
and it may include placeholders (e.g., ``xx'', ``someone'', ``something'').

\textbf{Task:} Based on this recalled line, generate $K$ possible original lyric expressions
for downstream retrieval.

\textbf{Requirements:}  Each candidate should be a natural and coherent lyric line. You may reference pronunciation, rhythm, imagery, or common lyric phrasing.

\textbf{Additional Constraints:} Literal overlap with the input is not required. Do not restrict yourself to word-by-word edits. Do not introduce clearly unrelated topics or scenarios.

\textbf{Notes:}  Words in the input may not appear in the original lyrics. Placeholders may be replaced with reasonable words or phrases. You do not need to determine which candidate is correct.

\textbf{Output:} One lyric per line. No numbering, explanations, or extra text.

\textbf{User recall:} \texttt{"\{lyric\}"}
\end{minipage}}
\caption{Prompt for lyrics-retrieval  hypothesis generation.}
\label{fig:lyrics_prompt}
\end{figure}

\noindent\textbf{Real-World BEIR Benchmarks.}
For BEIR datasets, we prompt the LLM to generate $K$ plausible \emph{original-intent hypotheses} that could have produced the observed query under imperfect recall, rather than to rewrite the query. For FIQA, the hypotheses express the same underlying financial information need while removing incidental narrative and avoiding any new assumptions. For SciFact, the hypotheses are alternative recalled statements of the same scientific claim. For NFCorpus, the hypotheses are alternative recalled questions preserving the same medical/nutrition information need.

\subsection{Retrievers}
\label{subsec:retrievers}

We evaluate our method on both sparse (lexical) and dense (embedding-based) retrievers to assess robustness across retrieval paradigms.

We use BM25 \citep{robertson1995okapi} as a strong lexical baseline. Given a query, BM25 ranks documents by term-level matching with TF--IDF-style weighting and document-length normalization.

We additionally evaluate two multilingual dense retrievers based on pretrained sentence encoders. For each dense retriever, we embed queries and documents independently and retrieve nearest neighbors by cosine similarity. Concretely, we precompute corpus embeddings and index them with FAISS \citep{douze2025faiss} using an inner-product index (\texttt{IndexFlatIP}); with $L_2$ normalization applied to both query and document embeddings, inner product search is equivalent to cosine similarity. 

The two embedding backbones we use  are:
(i) \texttt{sentence-transfor}

\noindent\texttt{mers/paraphrase-multilingual-MiniLM-L12-v2} \citep{reimers2019sentence,reimers2020making} (Dense 1), a lightweight multilingual encoder commonly used for semantic similarity, and
(ii) \texttt{Alibaba-NLP/gte-multilingual-base} (Dense 2) \citep{li2023towards}, a higher-capacity multilingual embedding model designed for general-purpose retrieval.
Both dense retrievers are used off-the-shelf without task-specific fine-tuning, so any gains can be attributed to QUARK rather than retriever adaptation.

\subsection{Simulation Results}
\label{subsec:simulation_results}

Table~\ref{tab:sim_all_alpha_makecell} reports lyrics-retrieval simulation results across three retrievers (BM25, Dense-1, Dense-2) and three query-faithfulness regimes (L1, L2, L3). For each retriever and regime, we compare the base retriever with its QUARK-enhanced counterpart. For QUARK, we sweep the aggregation parameter $\alpha$ over a fixed grid
$\{0.1, \ldots, 0.9,$ $0.91, \ldots, 0.99\}$ and report the best result for each setting. 
We use this best-$\alpha$ summary to characterize the attainable performance of QUARK (independent of retriever training), while a separate analysis in Section~\ref{sec:alpha_choice} studies sensitivity to $\alpha$ and provides practical guidance for selecting $\alpha$ without oracle access.


For BM25, QUARK yields consistent gains across all regimes. Improvements in MRR and nDCG@10 are statistically significant at all levels, including the hardest L3 regime. Notably, the optimal $\alpha$ decreases as query faithfulness degrades (from $\alpha=0.90$ at L1 to $\alpha=0.20$ at L3), indicating that heavier reliance on recovery hypotheses is beneficial when lexical mismatch becomes severe.

For Dense-1, QUARK provides limited benefit under L1, where the base retriever already performs strongly. Under L2, QUARK yields substantial and statistically significant improvements in both MRR and nDCG@10, with $\alpha=0.50$ balancing the original query and recovered hypotheses. Under L3, gains are smaller and not statistically significant, suggesting that semantic recovery becomes less effective when the observed query contains insufficient signal.

Dense-2 exhibits the strongest base performance across all regimes. As a result, QUARK yields only marginal improvements under L1 and L3. Nevertheless, under L2, QUARK achieves statistically significant gains in both MRR and nDCG@10, again demonstrating that recovery hypothesis aggregation is most effective in the intermediate regime where semantic drift is present but recoverable. The optimal $\alpha$ remains close to 1.0 for Dense-2, reflecting the robustness of the underlying dense representations.
\begin{table}[t]
\centering
\footnotesize
\setlength{\tabcolsep}{4pt}
\begin{tabular}{c c c c c c c}
\toprule
Retriever & Level & Method & R@1 & R@5 & R@10 & MRR / nDCG@10 \\
\midrule

\multirow{6}{*}{BM25}
& \multirow{2}{*}{L1}
& Base
& 0.817 & 0.942 & 0.958 & 0.876 / 0.897 \\
& 
& +QUARK
& \makecell{0.826 \\ \scriptsize($\alpha{=}0.90$)}
& \makecell{0.946 \\ \scriptsize($\alpha{=}0.90$)}
& \makecell{0.960 \\ \scriptsize($\alpha{=}0.90$)}
& \makecell{\underline{\textbf{0.883}} \\ \scriptsize($\alpha{=}0.90$)} /
  \makecell{\underline{\textbf{0.903}} \\ \scriptsize($\alpha{=}0.90$)} \\

& \multirow{2}{*}{L2}
& Base
& 0.548 & 0.710 & 0.753 & 0.619 / 0.652 \\
& 
& +QUARK
& \makecell{0.575 \\ \scriptsize($\alpha{=}0.80$)}
& \makecell{0.719 \\ \scriptsize($\alpha{=}0.80$)}
& \makecell{0.761 \\ \scriptsize($\alpha{=}0.80$)}
& \makecell{\underline{\textbf{0.640}} \\ \scriptsize($\alpha{=}0.80$)} /
  \makecell{\underline{\textbf{0.670}} \\ \scriptsize($\alpha{=}0.80$)} \\

& \multirow{2}{*}{L3}
& Base
& 0.082 & 0.156 & 0.193 & 0.118 / 0.136 \\
& 
& +QUARK
& \makecell{0.090 \\ \scriptsize($\alpha{=}0.20$)}
& \makecell{0.166 \\ \scriptsize($\alpha{=}0.20$)}
& \makecell{0.208 \\ \scriptsize($\alpha{=}0.20$)}
& \makecell{\underline{\textbf{0.125}} \\ \scriptsize($\alpha{=}0.20$)} /
  \makecell{\underline{\textbf{0.145}} \\ \scriptsize($\alpha{=}0.20$)} \\

\midrule

\multirow{6}{*}{Dense-1}
& \multirow{2}{*}{L1}
& Base
& 0.833 & 0.919 & 0.928 & 0.877 / 0.891 \\
& 
& +QUARK
& \makecell{0.833 \\ \scriptsize($\alpha{=}0.95$)}
& \makecell{0.920 \\ \scriptsize($\alpha{=}0.95$)}
& \makecell{0.931 \\ \scriptsize($\alpha{=}0.95$)}
& \makecell{0.877 \\ \scriptsize($\alpha{=}0.95$)} /
  \makecell{0.892 \\ \scriptsize($\alpha{=}0.95$)} \\

& \multirow{2}{*}{L2}
& Base
& 0.597 & 0.714 & 0.744 & 0.650 / 0.673 \\
& 
& +QUARK
& \makecell{0.633 \\ \scriptsize($\alpha{=}0.50$)}
& \makecell{0.749 \\ \scriptsize($\alpha{=}0.50$)}
& \makecell{0.795 \\ \scriptsize($\alpha{=}0.50$)}
& \makecell{\underline{\textbf{0.687}} \\ \scriptsize($\alpha{=}0.50$)} /
  \makecell{\underline{\textbf{0.713}} \\ \scriptsize($\alpha{=}0.50$)} \\

& \multirow{2}{*}{L3}
& Base
& 0.245 & 0.382 & 0.432 & 0.307 / 0.337 \\
& 
& +QUARK
& \makecell{0.247 \\ \scriptsize($\alpha{=}0.80$)}
& \makecell{0.382 \\ \scriptsize($\alpha{=}0.80$)}
& \makecell{0.434 \\ \scriptsize($\alpha{=}0.80$)}
& \makecell{0.307 \\ \scriptsize($\alpha{=}0.80$)} /
  \makecell{0.338 \\ \scriptsize($\alpha{=}0.80$)} \\

\midrule

\multirow{6}{*}{Dense-2}
& \multirow{2}{*}{L1}
& Base
& 0.908 & 0.982 & 0.987 & 0.946 / 0.957 \\
& 
& +QUARK
& \makecell{0.909 \\ \scriptsize($\alpha{=}0.99$)}
& \makecell{0.982 \\ \scriptsize($\alpha{=}0.99$)}
& \makecell{0.986 \\ \scriptsize($\alpha{=}0.99$)}
& \makecell{0.946 \\ \scriptsize($\alpha{=}0.99$)} /
  \makecell{0.957 \\ \scriptsize($\alpha{=}0.99$)} \\

& \multirow{2}{*}{L2}
& Base
& 0.793 & 0.921 & 0.939 & 0.850 / 0.873 \\
& 
& +QUARK
& \makecell{0.796 \\ \scriptsize($\alpha{=}0.95$)}
& \makecell{0.921 \\ \scriptsize($\alpha{=}0.95$)}
& \makecell{0.939 \\ \scriptsize($\alpha{=}0.95$)}
& \makecell{\underline{\textbf{0.853}} \\ \scriptsize($\alpha{=}0.95$)} /
  \makecell{\underline{\textbf{0.875}} \\ \scriptsize($\alpha{=}0.95$)} \\

& \multirow{2}{*}{L3}
& Base
& 0.392 & 0.565 & 0.627 & 0.469 / 0.508 \\
& 
& +QUARK
& \makecell{0.392 \\ \scriptsize($\alpha{=}0.99$)}
& \makecell{0.566 \\ \scriptsize($\alpha{=}0.99$)}
& \makecell{0.631 \\ \scriptsize($\alpha{=}0.99$)}
& \makecell{0.469 \\ \scriptsize($\alpha{=}0.99$)} /
  \makecell{0.509 \\ \scriptsize($\alpha{=}0.99$)} \\

\bottomrule
\end{tabular}
\caption{Lyrics-retrieval simulation results across retrievers and query-faithfulness regimes. For QUARK, the best performance over a fixed $\alpha$ grid is reported. Bold and underlined values indicate statistically significant improvements over the base retriever (two-sided paired $t$-test, $p<0.05$).}
\label{tab:sim_all_alpha_makecell}
\end{table}

Overall, the results demonstrate that QUARK provides a robust, training-free improvement across both lexical and dense retrievers, with the largest and most consistent gains arising under moderately non-faithful queries. These findings highlight the importance of aggregating multiple intent-preserving hypotheses when the observed query deviates from the user’s original intent, while preserving the base retriever’s strengths when the query remains reliable.

\subsection{Real-World Benchmark Results}
\label{subsec:real_results}

Table~\ref{tab:real_all} reports results on three BEIR benchmarks (FIQA, SciFact, and NFCorpus) using three retrievers. Across all retrievers and datasets, QUARK always improves ranking-sensitive metrics (MRR and nDCG@10), with many gains reaching statistical significance.

For BM25, QUARK yields clear and statistically significant improvements on all three datasets. 

For dense retrievers, QUARK continues to provide universal gains. Dense-1 benefits substantially from QUARK on all datasets, with statistically significant improvements in both MRR and nDCG@10. Dense-2, which exhibits the strongest base performance, shows more modest but still statistically significant gains on FIQA and SciFact, while improvements on NFCorpus are smaller and concentrated primarily in nDCG@10.

Overall, these results demonstrate that QUARK generalizes beyond controlled simulations to real-world retrieval benchmarks, improving ranking quality across both lexical and dense retrievers without retriever fine-tuning. The magnitude of improvement correlates with the degree of non-faithfulness in user queries, reinforcing QUARK’s role as an effective recovery mechanism for latent user intent.

\begin{table}[t]
\centering
\footnotesize
\setlength{\tabcolsep}{4pt}
\begin{tabular}{c c c c c c c}
\toprule
Retriever & Dataset & Method & R@1 & R@5 & R@10 & MRR / nDCG@10 \\
\midrule

\multirow{6}{*}{BM25}
& \multirow{2}{*}{FIQA}
& Base & 0.204 & 0.364 & 0.449 & 0.271 / 0.218 \\
&  & +QUARK
& \makecell{0.221 \\ \scriptsize($\alpha{=}0.60$)}
& 0.392
& 0.471
& \makecell{\underline{\textbf{0.290}} \\ \scriptsize($\alpha{=}0.60$)} /
  \makecell{\underline{\textbf{0.235}} \\ \scriptsize($\alpha{=}0.60$)} \\

& \multirow{2}{*}{SciFact}
& Base & 0.523 & 0.753 & 0.797 & 0.618 / 0.652 \\
&  & +QUARK
& \makecell{0.573 \\ \scriptsize($\alpha{=}0.70$)}
& 0.767
& 0.817
& \makecell{\underline{\textbf{0.658}} \\ \scriptsize($\alpha{=}0.70$)} /
  \makecell{\underline{\textbf{0.685}} \\ \scriptsize($\alpha{=}0.70$)} \\

& \multirow{2}{*}{NFCorpus}
& Base & 0.427 & 0.629 & 0.706 & 0.511 / 0.310 \\
&  & +QUARK
& \makecell{0.449 \\ \scriptsize($\alpha{=}0.70$)}
& 0.647
& 0.706
& \makecell{\underline{\textbf{0.531}} \\ \scriptsize($\alpha{=}0.70$)} /
  \makecell{\underline{\textbf{0.319}} \\ \scriptsize($\alpha{=}0.70$)} \\

\midrule

\multirow{6}{*}{Dense-1}
& \multirow{2}{*}{FIQA}
& Base & 0.201 & 0.347 & 0.420 & 0.262 / 0.203 \\
&  & +QUARK
& \makecell{0.198 \\ \scriptsize($\alpha{=}0.95$)}
& 0.363
& 0.432
& \makecell{\underline{\textbf{0.267}} \\ \scriptsize($\alpha{=}0.95$)} /
  \makecell{\underline{\textbf{0.209}} \\ \scriptsize($\alpha{=}0.95$)} \\

& \multirow{2}{*}{SciFact}
& Base & 0.393 & 0.550 & 0.633 & 0.455 / 0.484 \\
&  & +QUARK
& \makecell{0.433 \\ \scriptsize($\alpha{=}0.20$)}
& 0.583
& 0.667
& \makecell{\underline{\textbf{0.494}} \\ \scriptsize($\alpha{=}0.20$)} /
  \makecell{\underline{\textbf{0.520}} \\ \scriptsize($\alpha{=}0.20$)} \\

& \multirow{2}{*}{NFCorpus}
& Base & 0.325 & 0.514 & 0.585 & 0.407 / 0.236 \\
&  & +QUARK
& \makecell{0.372 \\ \scriptsize($\alpha{=}0.30$)}
& 0.567
& 0.635
& \makecell{\underline{\textbf{0.453}} \\ \scriptsize($\alpha{=}0.30$)} /
  \makecell{\underline{\textbf{0.267}} \\ \scriptsize($\alpha{=}0.30$)} \\

\midrule

\multirow{6}{*}{Dense-2}
& \multirow{2}{*}{FIQA}
& Base & 0.446 & 0.644 & 0.732 & 0.532 / 0.450 \\
&  & +QUARK
& \makecell{0.454 \\ \scriptsize($\alpha{=}0.70$)}
& 0.653
& 0.747
& \makecell{\underline{\textbf{0.540}} \\ \scriptsize($\alpha{=}0.70$)} /
  \makecell{\underline{\textbf{0.459}} \\ \scriptsize($\alpha{=}0.70$)} \\

& \multirow{2}{*}{SciFact}
& Base & 0.607 & 0.827 & 0.873 & 0.699 / 0.734 \\
&  & +QUARK
& \makecell{0.617 \\ \scriptsize($\alpha{=}0.90$)}
& 0.840
& 0.873
& \makecell{\underline{\textbf{0.706}} \\ \scriptsize($\alpha{=}0.90$)} /
  \makecell{\underline{\textbf{0.738}} \\ \scriptsize($\alpha{=}0.90$)} \\

& \multirow{2}{*}{NFCorpus}
& Base & 0.480 & 0.694 & 0.755 & 0.568 / 0.368 \\
&  & +QUARK
& \makecell{0.492 \\ \scriptsize($\alpha{=}0.80$)}
& 0.712
& 0.774
& 0.582 /
  \makecell{\underline{\textbf{0.376}} \\ \scriptsize($\alpha{=}0.80$)} \\

\bottomrule
\end{tabular}
\caption{Results on real-world BEIR benchmarks. For QUARK, the best-performing $\alpha$ selected via grid search is reported. Statistical significance is evaluated only for MRR and nDCG@10 using a two-sided paired $t$-test; bold underlined values indicate $p<0.05$.}
\label{tab:real_all}
\end{table}
\section{Sensitivity Analysis}
\label{sec:alpha_choice}
We study the sensitivity of QUARK to the aggregation parameter $\alpha$ by sweeping over the same fixed grid used throughout the paper,
$\{0.1,\ldots,0.9,0.91,\ldots,0.99\}$, and evaluating retrieval quality as a function of $\alpha$.
We visualize sensitivity using BM25 only. As a purely lexical retriever, BM25 is typically the most sensitive to the trade-off induced by $\alpha$---preserving exact query signal versus incorporating recovered hypotheses---and thus serves as a conservative, representative case.
Dense retrievers exhibit qualitatively similar but flatter trends, consistent with the relatively large optimal $\alpha$ values and modest performance differences in Tables~\ref{tab:sim_all_alpha_makecell} and~\ref{tab:real_all}; we therefore omit additional dense-retriever plots for brevity.

\begin{figure}[t]
    \centering
    \includegraphics[width=0.85\linewidth]{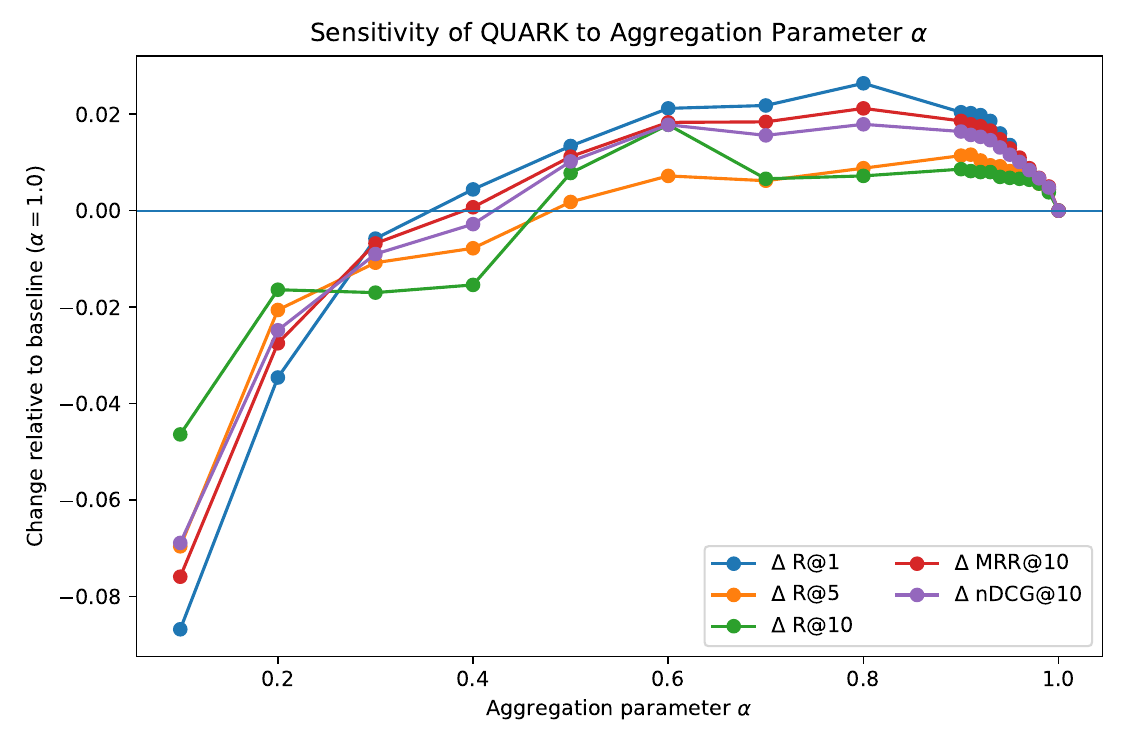}    \caption{Sensitivity of QUARK to $\alpha$ on the lyrics-retrieval simulation (L2), reported as changes relative to the baseline retriever ($\alpha=1.0$).}
    \label{fig:alpha_sensitivity_delta}
\end{figure}

\noindent\textbf{Lyrics-retrieval simulation.}
Figure~\ref{fig:alpha_sensitivity_delta} reports $\alpha$ sensitivity at the moderately non-faithful queries (L2), where the benefit--noise trade-off is most visible.
As $\alpha$ decreases from 1.0, performance improves smoothly, peaks at an intermediate range, and degrades when $\alpha$ becomes too small.

\begin{figure*}[t]
    \centering
    \includegraphics[width=\textwidth]{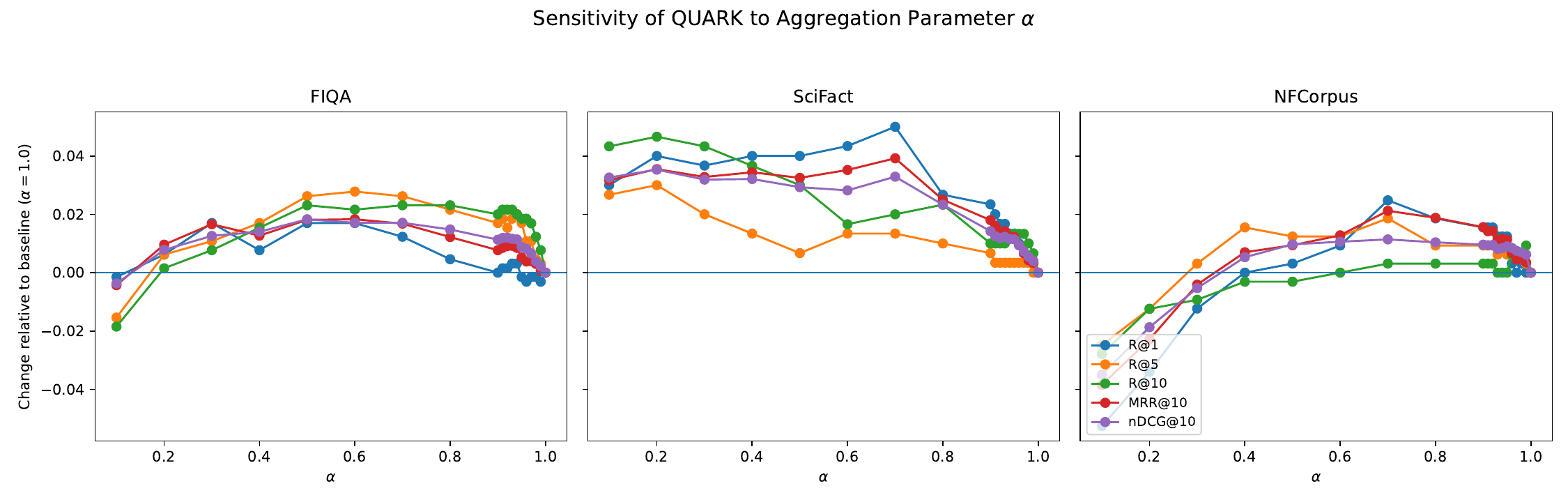}
    \Description{Three-panel line plots for FIQA, SciFact, and NFCorpus showing metric changes relative to baseline as $\alpha$ varies under BM25.}
    \caption{Sensitivity of QUARK to $\alpha$ on BEIR benchmarks with BM25, reported as changes relative to the baseline retriever ($\alpha=1.0$).}
    \label{fig:sensitivity-all-datasets}
\end{figure*}

\noindent\textbf{Real-world BEIR benchmarks.}
Figure~\ref{fig:sensitivity-all-datasets} shows the same analysis on FIQA, SciFact, and NFCorpus.
Across datasets, QUARK exhibits smooth and stable behavior: performance improves when moving from $\alpha=1.0$ to moderately smaller values, reaches a broad plateau, and gradually declines for very small $\alpha$.
For $\alpha$-selection, Empirically, $\alpha \in [0.6,0.9]$ provides a practical operating range for BM25 on real data without requiring fine-grained tuning.

\noindent{\textbf{Empirical guidence in selecting $\alpha$.}} Figures \ref{fig:alpha_sensitivity_delta}–\ref{fig:sensitivity-all-datasets} show a consistent inverted-U trend: performance drops sharply for small $\alpha$ (over-reliance on recovery hypotheses) but is stable and near-optimal over a broad high-
$\alpha$ plateau. In practice, fixing 
$\alpha\in[0.7,0.9]$ provides robust gains across datasets/metrics while smoothly reverting to the baseline as $\alpha\to 1$, requiring no oracle tuning.

\section{Ablation Studies}\label{sec:ablation studies}

\subsection{Effect of the Recovery Hypotheses}
\label{sec:ablation_k}

We first analyze the effect of the number of recovery hypotheses in a controlled simulation setting using lyrics retrieval under L2 queries. Table~\ref{tab:ablation-num-hypotheses} reports results with BM25 and QUARK, fixing $\alpha=0.8$. The results indicate that QUARK does not rely on a precise choice of $K$.

We next conduct the same ablation on real-world benchmarks (FIQA, SciFact, and NFCorpus), fixing $\alpha$ to the value used in the main experiments to isolate the effect of $K$. As shown in Table~\ref{tab:ablation_k_vertical}, adding recovery hypotheses can largely improve ranking performance over the baseline, while differences among $K\ge 1$ are modest. 

Overall, these results demonstrate that QUARK is robust to the number of recovery hypotheses once at least one hypothesis is used. Based on this robustness and to encourage hypothesis diversity without additional tuning, we adopt $K=5$ as a simple default in the main experiments.

\begin{table}[t]
\centering
\small
\setlength{\tabcolsep}{4pt}
\begin{tabular}{cccccc}
\toprule
$K$ 
& Recall@1 & Recall@5 & Recall@10 
& MRR@10 & nDCG@10 \\
\midrule
0 & 0.548 & 0.710 & 0.753 & 0.619 & 0.652 \\
1 & 0.576 & 0.720 & 0.759 & \underline{\textbf{0.639}} & \underline{\textbf{0.668}} \\
2 & 0.576 & 0.719 & 0.761 & \underline{\textbf{0.639}} & \underline{\textbf{0.669}} \\
3 & 0.577 & 0.718 & 0.761 & \underline{\textbf{0.640}} & \underline{\textbf{0.669}} \\
4 & 0.575 & 0.717 & 0.761 & \underline{\textbf{0.638}} & \underline{\textbf{0.668}} \\
5 & 0.575 & 0.720 & 0.761 & \underline{\textbf{0.640}} & \underline{\textbf{0.670}} \\
\bottomrule
\end{tabular}
\caption{Ablation on the number of recovery hypotheses for lyrics retrieval under L2 queries
(BM25, $\alpha=0.8$). Bold and underlined values indicate statistically significant improvements over the base retriever (two-sided paired $t$-test, $p<0.05$).}
\label{tab:ablation-num-hypotheses}
\end{table}
\begin{table}[t]
\centering
\small
\setlength{\tabcolsep}{6pt}
\begin{tabular}{c|ccccc}
\toprule
\multicolumn{6}{c}{\textbf{FIQA (BM25 + QUARK, $\alpha=0.6$)}} \\
\midrule
$K$
& Recall@1 & Recall@5 & Recall@10 & MRR@10 & nDCG@10 \\
\midrule
0 & 0.204 & 0.364 & 0.449 & 0.271 & 0.218 \\
1 & 0.207 & 0.367 & 0.451 & 0.276 & 0.221 \\
2 & 0.228 & 0.378 & 0.468 & \underline{\textbf{0.294}} & \underline{\textbf{0.236}} \\
3 & 0.221 & 0.387 & 0.469 & \underline{\textbf{0.291}} & \underline{\textbf{0.235}} \\
4 & 0.221 & 0.387 & 0.469 & 0.288 & \underline{\textbf{0.233}} \\
5 & 0.221 & 0.392 & 0.471 & \underline{\textbf{0.290}} & \underline{\textbf{0.235}} \\
\midrule
\multicolumn{6}{c}{\textbf{SciFact (BM25 + QUARK, $\alpha=0.7$)}} \\
\midrule
0 & 0.523 & 0.753 & 0.797 & \underline{\textbf{0.618}} & \underline{\textbf{0.652}} \\
1 & 0.563 & 0.763 & 0.823 & \underline{\textbf{0.652}} & \underline{\textbf{0.681}} \\
2 & 0.560 & 0.767 & 0.820 & \underline{\textbf{0.653}} & \underline{\textbf{0.682}} \\
3 & 0.573 & 0.767 & 0.817 & \underline{\textbf{0.660}} & \underline{\textbf{0.687}} \\
4 & 0.567 & 0.767 & 0.817 & \underline{\textbf{0.655}}& \underline{\textbf{0.683}} \\
5 & 0.573 & 0.767 & 0.817 & \underline{\textbf{0.658}} & \underline{\textbf{0.685}} \\
\midrule
\multicolumn{6}{c}{\textbf{NFCorpus (BM25 + QUARK, $\alpha=0.7$)}} \\
\midrule
0 & 0.427 & 0.629 & 0.706 & 0.511 & 0.310 \\
1 & 0.446 & 0.638 & 0.690 & 0.528 & 0.306 \\
2 & 0.440 & 0.650 & 0.703 & 0.524 & 0.315 \\
3 & 0.446 & 0.650 & 0.706 & \underline{\textbf{0.528}} & \underline{\textbf{0.320}} \\
4 & 0.437 & 0.644 & 0.712 & 0.525 & \underline{\textbf{0.319}} \\
5 & 0.449 & 0.647 & 0.706 & \underline{\textbf{0.531}} & \underline{\textbf{0.319}} \\
\bottomrule
\end{tabular}
\caption{Ablation on the number of recovery hypotheses ($K$) for QUARK with BM25 for FIQA, SciFact, and NFCorpus with fixed $\alpha$.
 Bold and underlined values indicate statistically significant improvements over the base retriever (two-sided paired $t$-test, $p<0.05$).}
\label{tab:ablation_k_vertical}
\end{table}

\begin{table}[t]
\centering
\small
\setlength{\tabcolsep}{6pt}
\begin{tabular}{c|ccccc}
\toprule
Aggregation
& R@1 & R@5 & R@10 & MRR@10 & nDCG@10 \\
\midrule
Median
& 0.338 & 0.494 & 0.560 & 0.406 & 0.443 \\
Mean
& 0.433 & 0.638 & 0.732 & 0.522 & 0.572 \\
Max
& 0.530 & 0.705 & \textbf{0.767} & 0.606 & 0.645 \\
\midrule
\textbf{QUARK}
& \textbf{0.575} & \textbf{0.719} & 0.761 & \textbf{0.640} & \textbf{0.670} \\
\bottomrule
\end{tabular}

\caption{Ablation on aggregation strategy for lyrics retrieval under L2 moderately non-faithful queries (BM25).
Unanchored strategies directly pool scores across the original query and recovery hypotheses using max, mean, or median.
QUARK uses query-anchored aggregation with fixed $\alpha$.}
\label{tab:ablation}
\end{table}

\subsection{Effect of Aggregation Strategy}
Instead of Query-Anchored aggregation, we ablate on other pooling ways, including Unanchored-Max pooling, Unanchored-Mean pooling, and Unanchored-Median pooling.
These unanchored strategies directly pool retrieval scores across the original query and all recovery hypotheses, treating them symmetrically without explicitly preserving the contribution of the original query.

As shown in Table~\ref{tab:ablation}, unanchored max pooling is the strongest among these baselines, while mean and median pooling suffer from substantial performance degradation due to score dilution.
Nevertheless, all unanchored variants underperform QUARK, demonstrating that explicitly anchoring aggregation to the original query is critical for robust retrieval under non-faithful queries.

Due to space constraints, we report aggregation ablations only in the controlled simulation setting. We observe similar trends on real-world benchmarks, and therefore use QUARK’s query-anchored aggregation with fixed $\alpha$ in all main experiments.

\section{Conclusion and Discussion}
We formalized retrieval under non‑faithful queries through a recall‑noise model and proposed QUARK, a simple yet effective, training‑free framework that anchors aggregation to the original query while incorporating evidence from recovery hypotheses. Experiments demonstrate consistent gains across retrievers and noise levels, with ablations confirming that anchored aggregation is crucial for preventing semantic drift. While QUARK is robust to hypothesis count and noise, its reliance on black‑box LLMs for hypothesis generation and a fixed anchoring parameter presents opportunities for refinement; future work includes learning-based adaptation of $\alpha$, lightweight fine‑tuning, and extension to conversational search.
\bibliographystyle{ACM-Reference-Format}
\bibliography{sample-base}

@inproceedings{zhang2020query,
  title={Query understanding via intent description generation},
  author={Zhang, Ruqing and Guo, Jiafeng and Fan, Yixing and Lan, Yanyan and Cheng, Xueqi},
  booktitle={Proceedings of the 29th ACM International Conference on Information \& Knowledge Management},
  pages={1823--1832},
  year={2020}
}

@article{azad2022improving,
  title={Improving query expansion using pseudo-relevant web knowledge for information retrieval},
  author={Azad, Hiteshwar Kumar and Deepak, Akshay and Chakraborty, Chinmay and Abhishek, Kumar},
  journal={Pattern Recognition Letters},
  volume={158},
  pages={148--156},
  year={2022},
  publisher={Elsevier}
}

@inproceedings{bodner1996knowledge,
  title={Knowledge-based approaches to query expansion in information retrieval},
  author={Bodner, Richard C and Song, Fei},
  booktitle={Conference of the Canadian Society for Computational Studies of Intelligence},
  pages={146--158},
  year={1996},
  organization={Springer}
}

@article{carpineto2012survey,
  title={A survey of automatic query expansion in information retrieval},
  author={Carpineto, Claudio and Romano, Giovanni},
  journal={Acm Computing Surveys (CSUR)},
  volume={44},
  number={1},
  pages={1--50},
  year={2012},
  publisher={ACM New York, NY, USA}
}

@inproceedings{liu2025robust,
  title={Robust information retrieval},
  author={Liu, Yu-An and Zhang, Ruqing and Guo, Jiafeng and de Rijke, Maarten},
  booktitle={Proceedings of the Eighteenth ACM International Conference on Web Search and Data Mining},
  pages={1008--1011},
  year={2025}
}

@article{behnamghader2024llm2vec,
  title={Llm2vec: Large language models are secretly powerful text encoders},
  author={BehnamGhader, Parishad and Adlakha, Vaibhav and Mosbach, Marius and Bahdanau, Dzmitry and Chapados, Nicolas and Reddy, Siva},
  journal={arXiv preprint arXiv:2404.05961},
  year={2024}
}

@article{krishna2023paraphrasing,
  title={Paraphrasing evades detectors of ai-generated text, but retrieval is an effective defense},
  author={Krishna, Kalpesh and Song, Yixiao and Karpinska, Marzena and Wieting, John and Iyyer, Mohit},
  journal={Advances in Neural Information Processing Systems},
  volume={36},
  pages={27469--27500},
  year={2023}
}

@inproceedings{khattab2020colbert,
  title={Colbert: Efficient and effective passage search via contextualized late interaction over bert},
  author={Khattab, Omar and Zaharia, Matei},
  booktitle={Proceedings of the 43rd International ACM SIGIR conference on research and development in Information Retrieval},
  pages={39--48},
  year={2020}
}

@article{xiong2020approximate,
  title={Approximate nearest neighbor negative contrastive learning for dense text retrieval},
  author={Xiong, Lee and Xiong, Chenyan and Li, Ye and Tang, Kwok-Fung and Liu, Jialin and Bennett, Paul and Ahmed, Junaid and Overwijk, Arnold},
  journal={arXiv preprint arXiv:2007.00808},
  year={2020}
}

@inproceedings{karpukhin2020dense,
  title={Dense Passage Retrieval for Open-Domain Question Answering.},
  author={Karpukhin, Vladimir and Oguz, Barlas and Min, Sewon and Lewis, Patrick SH and Wu, Ledell and Edunov, Sergey and Chen, Danqi and Yih, Wen-tau},
  booktitle={EMNLP (1)},
  pages={6769--6781},
  year={2020}
}

@article{salton1988term,
  title={Term-weighting approaches in automatic text retrieval},
  author={Salton, Gerard and Buckley, Christopher},
  journal={Information processing \& management},
  volume={24},
  number={5},
  pages={513--523},
  year={1988},
  publisher={Elsevier}
}

@book{robertson1995okapi,
  title={Okapi at TREC-3},
  author={Robertson, Stephen E and Walker, Steve and Jones, Susan and Hancock-Beaulieu, Micheline M and Gatford, Mike and others},
  year={1995},
  publisher={British Library Research and Development Department}
}

@article{belkin1980anomalous,
  title={Anomalous states of knowledge as a basis for information retrieval},
  author={Belkin, Nicholas J},
  journal={Canadian journal of information science},
  volume={5},
  number={1},
  pages={133--143},
  year={1980}
}

@article{furnas1987vocabulary,
  title={The vocabulary problem in human-system communication},
  author={Furnas, George W. and Landauer, Thomas K. and Gomez, Louis M. and Dumais, Susan T.},
  journal={Communications of the ACM},
  volume={30},
  number={11},
  pages={964--971},
  year={1987},
  publisher={ACM New York, NY, USA}
}

@inproceedings{jansen1998real,
  title={Real life information retrieval: A study of user queries on the web},
  author={Jansen, Bernard J and Spink, Amanda and Bateman, Judy and Saracevic, Tefko},
  booktitle={Acm sigir forum},
  volume={32},
  number={1},
  pages={5--17},
  year={1998},
  organization={ACM New York, NY, USA}
}

@article{krovetz1992lexical,
  title={Lexical ambiguity and information retrieval},
  author={Krovetz, Robert and Croft, W Bruce},
  journal={ACM Transactions on Information Systems (TOIS)},
  volume={10},
  number={2},
  pages={115--141},
  year={1992},
  publisher={ACM New York, NY, USA}
}

@article{jansen2009patterns,
  title={Patterns of query reformulation during web searching},
  author={Jansen, Bernard J and Booth, Danielle L and Spink, Amanda},
  journal={Journal of the american society for information science and technology},
  volume={60},
  number={7},
  pages={1358--1371},
  year={2009},
  publisher={Wiley Online Library}
}

@inproceedings{chen2021towards,
  title={Towards a better understanding of query reformulation behavior in web search},
  author={Chen, Jia and Mao, Jiaxin and Liu, Yiqun and Zhang, Fan and Zhang, Min and Ma, Shaoping},
  booktitle={Proceedings of the web conference 2021},
  pages={743--755},
  year={2021}
}

@inproceedings{liu2024perturbation,
  title={Perturbation-invariant adversarial training for neural ranking models: improving the effectiveness-robustness trade-off},
  author={Liu, Yu-An and Zhang, Ruqing and Zhang, Mingkun and Chen, Wei and de Rijke, Maarten and Guo, Jiafeng and Cheng, Xueqi},
  booktitle={Proceedings of the AAAI Conference on Artificial Intelligence},
  volume={38},
  number={8},
  pages={8832--8840},
  year={2024}
}

@article{elgohary2019can,
  title={Can you unpack that? learning to rewrite questions-in-context},
  author={Elgohary, Ahmed and Peskov, Denis and Boyd-Graber, Jordan},
  journal={Can You Unpack That? Learning to Rewrite Questions-in-Context},
  year={2019}
}

@article{thakur2021beir,
  title={Beir: A heterogenous benchmark for zero-shot evaluation of information retrieval models},
  author={Thakur, Nandan and Reimers, Nils and R{\"u}ckl{\'e}, Andreas and Srivastava, Abhishek and Gurevych, Iryna},
  journal={arXiv preprint arXiv:2104.08663},
  year={2021}
}

@article{azad2019query,
  title={Query expansion techniques for information retrieval: a survey},
  author={Azad, Hiteshwar Kumar and Deepak, Akshay},
  journal={Information Processing \& Management},
  volume={56},
  number={5},
  pages={1698--1735},
  year={2019},
  publisher={Elsevier}
}

@inproceedings{qian2022explicit,
  title={Explicit query rewriting for conversational dense retrieval},
  author={Qian, Hongjin and Dou, Zhicheng},
  booktitle={Proceedings of the 2022 Conference on Empirical Methods in Natural Language Processing},
  pages={4725--4737},
  year={2022}
}

@String{Computing = "Computing" }

@String{Springer = "Springer-Verlag" }

@inproceedings{sidiropoulos2022analysing,
  title={Analysing the robustness of dual encoders for dense retrieval against misspellings},
  author={Sidiropoulos, Georgios and Kanoulas, Evangelos},
  booktitle={Proceedings of the 45th International ACM SIGIR Conference on Research and Development in Information Retrieval},
  pages={2132--2136},
  year={2022}
}

@inproceedings{aliannejadi2019asking,
  title={Asking clarifying questions in open-domain information-seeking conversations},
  author={Aliannejadi, Mohammad and Zamani, Hamed and Crestani, Fabio and Croft, W Bruce},
  booktitle={Proceedings of the 42nd international acm sigir conference on research and development in information retrieval},
  pages={475--484},
  year={2019}
}

@inproceedings{ng1998phonetic,
  title={Phonetic recognition for spoken document retrieval},
  author={Ng, Kenney and Zue, Victor W},
  booktitle={Proceedings of the 1998 IEEE International Conference on Acoustics, Speech and Signal Processing, ICASSP'98 (Cat. No. 98CH36181)},
  volume={1},
  pages={325--328},
  year={1998},
  organization={IEEE}
}

@article{zhou2014iterative,
  title={Iterative refinement methods for enhanced information retrieval},
  author={Zhou, Dong and Truran, Mark and Liu, Jianxun and Li, Wei and Jones, Gareth},
  journal={International journal of intelligent systems},
  volume={29},
  number={4},
  pages={341--364},
  year={2014},
  publisher={Wiley Online Library}
}

@inproceedings{chen2018phonetic,
  title={Phonetic-and-semantic embedding of spoken words with applications in spoken content retrieval},
  author={Chen, Yi-Chen and Huang, Sung-Feng and Shen, Chia-Hao and Lee, Hung-Yi and Lee, Lin-Shan},
  booktitle={2018 IEEE Spoken Language Technology Workshop (SLT)},
  pages={941--948},
  year={2018},
  organization={IEEE}
}

@inproceedings{jang2024itercqr,
  title={IterCQR: Iterative conversational query reformulation with retrieval guidance},
  author={Jang, Yunah and Lee, Kang-il and Bae, Hyunkyung and Lee, Hwanhee and Jung, Kyomin},
  booktitle={Proceedings of the 2024 Conference of the North American Chapter of the Association for Computational Linguistics: Human Language Technologies (Volume 1: Long Papers)},
  pages={8121--8138},
  year={2024}
}

@inproceedings{zhou2023towards,
  title={Towards robust ranker for text retrieval},
  author={Zhou, Yucheng and Shen, Tao and Geng, Xiubo and Tao, Chongyang and Xu, Can and Long, Guodong and Jiao, Binxing and Jiang, Daxin},
  booktitle={Findings of the Association for Computational Linguistics: ACL 2023},
  pages={5387--5401},
  year={2023}
}

@inproceedings{agarwal2023towards,
  title={Towards effective paraphrasing for information disguise},
  author={Agarwal, Anmol and Gupta, Shrey and Bonagiri, Vamshi and Gaur, Manas and Reagle, Joseph and Kumaraguru, Ponnurangam},
  booktitle={European Conference on Information Retrieval},
  pages={331--340},
  year={2023},
  organization={Springer}
}

@inproceedings{gan2019improving,
  title={Improving the robustness of question answering systems to question paraphrasing},
  author={Gan, Wee Chung and Ng, Hwee Tou},
  booktitle={Proceedings of the 57th annual meeting of the association for computational linguistics},
  pages={6065--6075},
  year={2019}
}

@article{li2023towards,
  title={Towards general text embeddings with multi-stage contrastive learning},
  author={Li, Zehan and Zhang, Xin and Zhang, Yanzhao and Long, Dingkun and Xie, Pengjun and Zhang, Meishan},
  journal={arXiv preprint arXiv:2308.03281},
  year={2023}
}

@article{reimers2020making,
  title={Making monolingual sentence embeddings multilingual using knowledge distillation},
  author={Reimers, Nils and Gurevych, Iryna},
  journal={arXiv preprint arXiv:2004.09813},
  year={2020}
}

@article{reimers2019sentence,
  title={Sentence-bert: Sentence embeddings using siamese bert-networks},
  author={Reimers, Nils and Gurevych, Iryna},
  journal={arXiv preprint arXiv:1908.10084},
  year={2019}
}

@article{liu2024query,
  title={Query rewriting via large language models},
  author={Liu, Jie and Mozafari, Barzan},
  journal={arXiv preprint arXiv:2403.09060},
  year={2024}
}

@article{douze2025faiss,
  title={The faiss library},
  author={Douze, Matthijs and Guzhva, Alexandr and Deng, Chengqi and Johnson, Jeff and Szilvasy, Gergely and Mazar{\'e}, Pierre-Emmanuel and Lomeli, Maria and Hosseini, Lucas and J{\'e}gou, Herv{\'e}},
  journal={IEEE Transactions on Big Data},
  year={2025},
  publisher={IEEE}
}

@misc{gaussic_chinese_lyric_corpus,
  author       = {gaussic},
  title        = {Chinese Lyric Corpus},
  howpublished = {GitHub repository},
  year         = {2020},
  url          = {https://github.com/gaussic/Chinese-Lyric-Corpus},
  note         = {Accessed: 2026-01-23}
}

@article{yang2025qwen3,
  title={Qwen3 technical report},
  author={Yang, An and Li, Anfeng and Yang, Baosong and Zhang, Beichen and Hui, Binyuan and Zheng, Bo and Yu, Bowen and Gao, Chang and Huang, Chengen and Lv, Chenxu and others},
  journal={arXiv preprint arXiv:2505.09388},
  year={2025}
}

@inproceedings{kong2023sparseembed,
  title={Sparseembed: Learning sparse lexical representations with contextual embeddings for retrieval},
  author={Kong, Weize and Dudek, Jeffrey M and Li, Cheng and Zhang, Mingyang and Bendersky, Michael},
  booktitle={Proceedings of the 46th International ACM SIGIR conference on research and development in information retrieval},
  pages={2399--2403},
  year={2023}
}

@inproceedings{song2025sparse,
  title={Sparse and Dense Retrievers Learn Better Together: Joint Sparse-Dense Optimization for Text-Image Retrieval},
  author={Song, Jonghyun and Lee, Youngjune and Cho, Gyu-Hwung and Song, Ilhyeon and Kim, Saehun and Jo, Yohan},
  booktitle={Proceedings of the 34th ACM International Conference on Information and Knowledge Management},
  pages={5268--5272},
  year={2025}
}

@inproceedings{cormack2009reciprocal,
  title={Reciprocal rank fusion outperforms condorcet and individual rank learning methods},
  author={Cormack, Gordon V and Clarke, Charles LA and Buettcher, Stefan},
  booktitle={Proceedings of the 32nd international ACM SIGIR conference on Research and development in information retrieval},
  pages={758--759},
  year={2009}
}

@inproceedings{fox1993combining,
  title={Combining evidence from multiple searches},
  author={Fox, Edward A and Koushik, M Prabhakar and Shaw, Joseph and Modlin, Russell and Rao, Durgesh and others},
  booktitle={The first text retrieval conference (TREC-1)},
  pages={319--328},
  year={1993}
}

@article{sun2024r,
  title={R-bot: An llm-based query rewrite system},
  author={Sun, Zhaoyan and Zhou, Xuanhe and Li, Guoliang and Yu, Xiang and Feng, Jianhua and Zhang, Yong},
  journal={arXiv preprint arXiv:2412.01661},
  year={2024}
}

@article{wang2023generative,
  title={Generative query reformulation for effective adhoc search},
  author={Wang, Xiao and MacAvaney, Sean and Macdonald, Craig and Ounis, Iadh},
  journal={arXiv preprint arXiv:2308.00415},
  year={2023}
}

@article{zhu2025large,
  title={Large language models for information retrieval: A survey},
  author={Zhu, Yutao and Yuan, Huaying and Wang, Shuting and Liu, Jiongnan and Liu, Wenhan and Deng, Chenlong and Chen, Haonan and Liu, Zheng and Dou, Zhicheng and Wen, Ji-Rong},
  journal={ACM Transactions on Information Systems},
  volume={44},
  number={1},
  pages={1--54},
  year={2025},
  publisher={ACM New York, NY}
}

@article{wang2024utilizing,
  title={Utilizing bert for information retrieval: Survey, applications, resources, and challenges},
  author={Wang, Jiajia and Huang, Jimmy Xiangji and Tu, Xinhui and Wang, Junmei and Huang, Angela Jennifer and Laskar, Md Tahmid Rahman and Bhuiyan, Amran},
  journal={ACM Computing Surveys},
  volume={56},
  number={7},
  pages={1--33},
  year={2024},
  publisher={ACM New York, NY}
}

@article{nogueira2019multi,
  title={Multi-stage document ranking with BERT},
  author={Nogueira, Rodrigo and Yang, Wei and Cho, Kyunghyun and Lin, Jimmy},
  journal={arXiv preprint arXiv:1910.14424},
  year={2019}
}

@inproceedings{afuan2019study,
  title={A study: query expansion methods in information retrieval},
  author={Afuan, Lasmedi and Ashari, Ahmad and Suyanto, Yohanes},
  booktitle={Journal of Physics: Conference Series},
  volume={1367},
  number={1},
  pages={012001},
  year={2019},
  organization={IOP Publishing}
}

@ArtifactSoftware{R,
    title = {R: A Language and Environment for Statistical Computing},
    author = {{R Core Team}},
    organization = {R Foundation for Statistical Computing},
    address = {Vienna, Austria},
    year = {2019},
    url = {https://www.R-project.org/},
}










\end{document}